\documentclass[letterpaper]{article}

\usepackage{aaai21} 
  \usepackage{times} 
  \usepackage{helvet} 
  \usepackage{courier} 
  \usepackage[hyphens]{url} 
  \usepackage{graphicx} 
  \usepackage{graphicx}  
  \usepackage{natbib} 
  \usepackage{caption} 
\usepackage[utf8]{inputenc} 
\usepackage[T1]{fontenc}    
\usepackage{url}            
\usepackage{booktabs}       
\usepackage{amsfonts}       
\usepackage{nicefrac}       
\usepackage{microtype}      
\usepackage{amsthm}
\usepackage{amsmath}
\usepackage{amssymb}
\usepackage{mathrsfs}
\usepackage{subcaption}
\usepackage[ruled, vlined]{algorithm2e}
\newtheorem*{lemma}{Lemma}
\newtheorem*{proposition}{Proposition}

\usepackage[switch]{lineno}

\title{Explaining Deep Reinforcement Learning Agents In The Atari Domain through a Surrogate Model}

\author{%
  Alexander Sieusahai, Matthew Guzdial\\
}

\affiliations{
    Computing Science Department, Alberta Machine Intelligence Institute (Amii) \\
    University of Alberta \\
    asieusah@ualberta.ca, guzdial@ualberta.ca
}

\begin{document}


\maketitle
\begin{abstract}
One major barrier to applications of deep Reinforcement Learning (RL) both inside and outside of games is the lack of explainability. 
In this paper, we describe a lightweight and effective method to derive explanations for deep RL agents, which we evaluate in the Atari domain. 
Our method relies on a transformation of the pixel-based input of the RL agent to an interpretable, percept-like input representation.
We then train a surrogate model, which is itself interpretable, to replicate the behavior of the target, deep RL agent.
Our experiments demonstrate that we can learn an effective surrogate that accurately approximates the underlying decision making of a target agent on a suite of Atari games.
\end{abstract}

\section{Introduction}

In the field of reinforcement learning (RL), there have been recent strides in domains thought to be previously intractable, such as playing Atari games \cite{mnih2013playing}. 
However, these agents are largely uninterpretable and opaque (commonly referred to as black boxes \cite{10.1145/3236009}).
With the right to explain imposed by GDPR \cite{EUdataregulations2018}, real world applications of RL could be hampered as there is no consistent way to derive accurate, interpretable explanations for RL agent actions.
This is of particular interest in applied fields such as the medical \cite{holzinger2017need} and judicial \cite{rudin2018optimized} domains, where a model would need to provide regular explanations comprehensible to a layperson. 

There have been prior methods developed to derive explanations in computer vision applications.
For example, saliency maps, where pixels important to a model's decision are highlighted and displayed to the user \cite{simonyan2013deep}, and t-SNE \cite{maaten2008visualizing}, that visualizes images that seem similar to the current image according to the model. 
These approaches highlight important features in making a decision, but are derived from uninterpretable models and do not explain a model's decision making.
Model agnostic methods exist \cite{ribeiro2016modelagnostic}, such as SHAP \cite{lundberg2016unexpected}, which describes the impact each individual feature makes on the deviation of the given prediction from the mean prediction. 
However, they rely on the feature set itself being interpretable, which is untrue with models that utilize raw pixels as input. 
If we had an accurate, interpretable model with an appropriate feature set there would be benefits further outside of offering explanations to laypeople.
For example, an expert could develop a mental model of the decision making of the agent, driving improvements in the algorithm that created the agent \cite{hayes2017improving}.

In this paper, we show that a simple, but surprisingly accurate explainable model can be learned that approximates the decision making of a deep reinforcement learning agent in the Atari domain. 
Our approach depends on a state transformation of the raw pixel input to a symbolic, interpretable representation. 
We demonstrate the proposed state transformation leads to an equivalent representation via proving it is a bijection of the original state space. 
Lastly, we show that, on a set of Atari games with simple sprites and animations, the behavior of deep reinforcement learning agents can be captured by a decision tree, an interpretable classifier, trained on this symbolic feature representation.
In this way the symbolic surrogate decision tree and deep neural network agent create a neuro-symbolic, interpretable system.

\section{Background}
\label{gen_inst}



Deep Q Networks (DQN) \cite{mnih2013playing} were the first successful demonstration of reinforcement learning in the Atari domain. 
A DQN consists of a deep approximator of the Q function \cite{10.5555/3312046}, and utilizes the Q-learning algorithm \cite{watkins}. 
A myriad of improvements, such as Double DQN \cite{hasselt2015deep}, Prioritized Experience Replay \cite{schaul2015prioritized} and the Dueling architecture \cite{wang2015dueling} were used in tandem to produce the Rainbow architecture, which combines these improvements in an agent that is much more sample efficient than prior DQN based agents \cite{hessel2017rainbow}. 
We use Rainbow as our testbed agent.

Explainable AI (XAI) \cite{molnar2019,sado2020explainable}, refers to a class of methods to create human-interpretable explanations of a model's behavior \cite{ehsan2017rationalization}. 
By understanding how the model makes its decisions, the human can then make informed decisions on how to accommodate or address the behavior of the model.
For decision making augmented with models, it is integral that the model is able to communicate its decision making in a human understandable manner, which often means a symbolic feature representation \cite{goebel2018explainable}.


Juozapaitis et al. described an XAI approach that incorporates the target RL agent by decomposing the rewards and Q values \cite{10.5555/3312046} into a vector representation, where each reward source is one component in the vector \cite{explainablerewarddecomp}.
One can then formulate the agent action given a certain state in terms of how the agent makes the tradeoff between different reward sources. 
In many Atari games, there is usually one main source of reward, and occasionally some large source of reward (such as the UFO in \textit{Space Invaders}). 
Often, the agent is acting without the presence of that large source of reward, leading to a lack of explanation for some actions utilizing this method. 
Additionally, this approach does not take into account the state of the environment at all, consequently giving no explanation with respect to the current state. 
For example, in \textit{Breakout}, we wouldn't be able to explain the action with respect to the paddle (the player controlled entity) and the ball (the main interactable entity), since the only reward source is from hitting blocks. 
In our work we focus on the game entities present in a given observation, augmented with position and velocity, as building blocks for explanations of the agent.

One intuitive approach would be to learn a structural causal model of the agent and explaining the actions of the agent via counterfactuals. 
Madumal et al. \cite{madumal2019explainable} take this route, and focus on answering why (or why not) an agent decides to take an action given the current state. 
This approach generates explanations that both satisfied human participants and increased the trust of the human participants with respect to the agent. 
However, the surrogate tends to be a relatively weak fit for the agent, resulting in less accurate explanations of the agent. 
This is a problem in mission critical systems, or systems for which an accurate mental model of the agent must be understood by the user, such as self driving cars \cite{shalevshwartz2016safe}. 
We show our approach to be highly accurate and that the explainable surrogate model captures the underlying decision making of the target RL agent.
However, we anticipate that Madumal et al.'s results should generalize to our approach, in terms of our similar explanations increasing user satisfaction and trust.

Saliency maps \cite{simonyan2013deep} are widely purported to be a strong way of understanding what portions of an image are responsible for the model output. 
Since Atari agents use the pixels of the frame as a state representation, it seems reasonable to utilize saliency maps in order to explain what influenced an agent in its decision making process. 
Greydanus et al. \cite{greydanus2018visualizing} explore the utilization of saliency maps in order to explain to humans what portions of an image an agent utilized in order to make its decisions as a way of garnering trust.
However, this approach does not scale to large amounts of data due to the limitation of examining the decision making process of an agent frame by frame, requiring human inspection.

\section{The Surrogate Model}
\label{headings}

Drawing on \cite{doshivelez2017rigorous}, we define interpretability as the ability to explain or present in understandable terms to a human. 
The problem we target in this paper is to be able to interpret deep reinforcement learning agents.
We focus on the Atari domain \cite{ALE}, in this initial exploration of our approach, given its popularity as a deep RL domain.
We use a surrogate model trained on interpretable features to predict what action the agent will take in a given state. 
We can then use this interpretable surrogate model to explain the agent, as a proxy for directly explaining the agent.
The intuition is that a sufficiently strong surrogate model will capture the agent's decision making with interpretable features, which can then be communicated to a human user.

In order for the surrogate model to be useful for interpretation, we must transform the frame (the raw pixels from an Atari game) into these interpretable features, as utilizing the $(x, y)$ raw pixels as features directly would make meaningful interpretation impossible. 
One alternative would be to have a human tag each frame with interpretable information, however this would be prohibitively time consuming.
We outline a simple and effective automated approach that transforms the raw pixels into a feature set that is both interpretable and rich enough to create a powerful surrogate model. 
Additionally, we describe the data gathering process for the training dataset of the surrogate model such that it can learn a strong approximation of the decision making process of the agent.

\subsection{Sprites As Symbolic Features}
If one were to ask a human expert for an explanation of their \textit{Breakout} policy, they might answer using the concept of the ball and the paddle. In particular, they might describe their policy in terms of the locations of the ball and paddle, the locations of all of the blocks, and the current velocity of the ball, as these features directly impact the score of the game. 

Inspired by this intuition, we chose our interpretable features as the location, velocity, and appearance of game entities. 
First, we automatically identify the different entities. 
To do this, we break each frame into groups of pixels to represent each game entity. 
We refer to these groupings as \textit{sprites} as they represent individual sprites of the game. These sprites have position, and their velocity can be obtained using the changes in sprite position. 
In contrast to the three dimensional tensor representing the $x$ position, $y$ position and the channel, our feature space consists of the set of sprites, their locations, and their velocities. 
This allows us to interpret the model utilizing the importance of sprites rather than the importance of pixels. 

\begin{algorithm}[t]
\SetAlgoLined
 Input: Frame $F \in P$, Neighbor function $N : P \to \mathbb{P} (P)$\;
 Initialise $C : \mathcal{C} \to \mathbb{N}$ such that $C(c) = 0 \forall c \in \mathcal{C}$\;
 Initialise $V = \emptyset, O = \emptyset$\;
 
 \For{$x, y, c \in F$}{
     $C(c) \leftarrow C(c) + 1$\;
 }
 $B \leftarrow argmax_{c \in \mathcal{C}} C(c)$\;
 
 \For{$p \in F$}{
  $x, y, c \leftarrow p$\;
  \eIf{$c \equiv B$ or $p \in V$}{
   continue\;
   }{
    $o \leftarrow DFS(p, V)$\;  
    $V \leftarrow V \cup o$\;  
    $O \leftarrow O \cup \{o\}$\;
  }
 }
 \Return $O$\;
 \caption{Greedy pixel-wise sprite identification}
 \label{algo}
\end{algorithm}

\subsection{Transforming Frames Into Features}

Given a frame, we need to automatically convert it into the feature space described above. 
To identify sprites, we need to first identify the background color, to avoid interpreting the background itself as a sprite. 
We do this by relying on a heuristic of the most common pixel color in a given frame. 
With the background color identified, we run a depth first search through every unvisited pixel that is not the background color. 
We consider neighboring pixels to be neighbors in the search space if they have the same pixel color, so that the set of explored frontiers forms a partition over the pixels. 
In order to increase generalization between similar sprites and to match the feature generation process for Rainbow \cite{hessel2017rainbow}, this process is run on a downsampled version of the image (downsampled from $(210, 160)$ to $(105, 80)$). 
In this way we can identify the sprites of the game without human input.
See Algorithm \ref{algo} for pseudocode of this procedure.
We refer to this algorithm as ``greedy pixel-wise sprite identification". 

Note that we assume that sprites in an Atari game have a uniform pixel color, and the entire sprite is structured so that every pixel in the sprite is adjacent to at least one other pixel in the sprite. 
As our results demonstrate, this approach is still viable for games where sprites have multiple segments of pixel colors (such as in \textit{Demon Attack}), as each segment of the larger sprite is highly correlated.
However, we anticipate that this naive method will need to be extended for domains with a broader palette of potential pixel colors and where individual pixels may have unique colors from all surrounding pixels, such as \textit{DOTA2} or \textit{Starcraft}.

\begin{figure}
  \includegraphics[width=4.1cm]{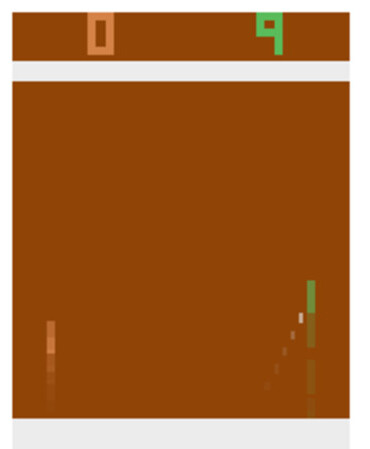}
  \includegraphics[width=4cm]{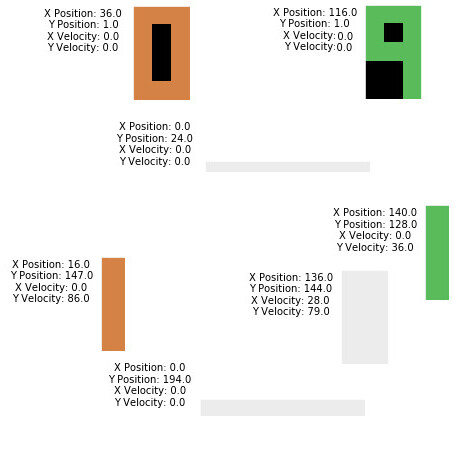}
  \centering
  \caption{A visual demonstration of the output features after the state transformation for a frame of \textit{Pong}. The left image is the last 5 frames smoothed to show movement, and the sprites and their corresponding features are shown to the right in their relative positions.}
  \label{pong}
\end{figure}

Now that we have the frame grouped into sprites, we need to transform this grouping into features. 
We designate the lowest and farthest left pixel coordinates of each sprite to be its $(x, y)$ coordinates. 
If multiple sprites with the same pixels are present, such as multiple aliens in Space Invaders, we sort them by their $(x, y)$ coordinates. 
Then, for a given sprite returned by this procedure, we have its pixels, $x$ coordinate, $y$ coordinate and its relative position. 
To find the velocity we treat the data as if it's a time series and take the difference of the data shifted back by one timestep and the current data.
See Figure \ref{pong} for an example of our representation.

\subsection{Analysis of Algorithm 1}

We provide a proof that our proposed state transformation is a bijection of the raw pixel inputs, to show concretely that the feature space is rich enough to train an accurate surrogate of the agent. 
Let $P = \mathbb{P} (\mathcal{X} \times \mathcal{Y} \times \mathcal{C})$ represent the set of all possible frames, where $\mathcal{X}$ and $\mathcal{Y}$ are the set of possible x, y coordinates respectively, $\mathcal{C}$ is the set of all possible RGB colors, and $\mathbb{P}(X)$ is the power set of $X$. Let $t : P \to Q \subset P$ be the state transformation function.

\begin{proposition}
The state transformation function $t$ separates the argument into a partition. That is, for any frame $F$, $\cup_{u \in t(F)} u = F$.
\end{proposition}

\textbf{Proof.} Let $F \in P$ be arbitrary. From Algorithm \ref{algo}, we have that for an arbitrary pixel $p \in F$, it must either belong to an object generated by the algorithm or a part of the background. Suppose that $p \in o_0$ and $p \in o_1$, for some $o_0, o_1 \in t(X)$. We then have that $p$ was visited multiple times in Algorithm \ref{algo}, which violates the loop invariant for the outermost loop. So, we have that for arbitrary $p \in X$, there exists $o$ in $t(X)$ such that $p \in o$. 
Additionally, suppose that there exists some $p \in F$ such that for all $o \in t(F)$, $p \not \in o$. We then have that $p$ must have not been visited in Algorithm \ref{algo}, which is impossible as we execute the depth first search on every $p \in X$ such that it hasn't been visited previously due to some other $p' \in F$. So, we have that $t(F)$ is a partition of $F$.

\begin{lemma}
The state transformation function $t$ is a bijection.
\end{lemma}

\textbf{Proof.} First, we show that $t$ is an injection. Let $X, Y \in P$ such that $t(X) = t(Y)$. Since $t(X) = t(Y) = O \subset \mathbb{P}(P)$, we have that for all $o \in O$ generated by Algorithm \ref{algo}, $o \subset X$ and $o \subset Y$. Since $\cup_{p \in o} p \in O$, we have that $X = Y$. Next, we show that $t$ is a surjection. Let $O$ be in the range of $t$. Since we have that $t$ forms a partition of the argument, we have that $\cup_{o \in O} o = X$ implies that $t(X) = O$. So, we have that $t$ is a bijection.

\subsection{Obtaining a Surrogate Model For the Agent}

Our environment is a Markov decision process $\mathcal{M} = (\mathcal{S, A}, P, r, \rho_0, \gamma)$ \cite{10.5555/3312046}.
There are two distinct cases that we would like to examine in our evaluation, in order to encourage diversity in our RL agent's behavior; \textit{noop starts} and \textit{sticky actions alongside noop starts}.
For \textit{noop starts}, we modify $\rho_0$ by executing $k$ noop actions, to diversify the distribution of starting states \cite{mnih2015human}.
For the case with \textit{sticky actions} \cite{machado2017revisiting}, the last selected action is used in place of the agent action with probability $\zeta=0.25$, as recommended in \cite{machado2017revisiting}, alongside the aforementioned noop starts modification.
This gives us more distinct trajectories and forces the agent to learn a more general policy, which we draw on for our evaluation.
In order to train a surrogate model, trajectories must be sampled; we employ a sampling procedure for obtaining trajectories corresponding to noop starts $k \to v$. 
For each noop start in $[k, k+1, ..., v]$, begin by sampling $s_0$ from $\rho_0$. 
Until termination of the episode, sample $a_t$ according to $\pi(\cdot | s_t)$, and obtain the new state and reward $s_{t+1}, r_{t+1}$, all while recording the mapping between $t(s_t)$ and $a_t$.
This allows us to divide our trajectories into training and test sets.
In the sticky actions case, the state features are augmented with the last action taken in order to expose this information to the surrogate model, which the agent has knowledge of implicitly via the state representation in \cite{mnih2013playing}.
This gives us a dataset $\mathcal{D}_k$, where $k$ is the amount of noop starts. 
Once the desired amount of trajectories are obtained, we train a classifier to approximate the mapping of $t(s_t) \to a_t$, using $\cup_{k \in K} \mathcal{D}_k$, where $K$ is the set of all noop starts used to generate the trajectories.
This is reminiscent of imitation learning \cite{imitation}, though in this case we want to approximate the agent's behavior in a symbolic representation, rather than a reward function. 
In order for the surrogate model to be interpretable, a model of desired interpretable ability should be selected. 
In particular, a decision tree has strong predictive power \cite{steinberg2009cart} while still achieving interpretability when trained on interpretable features. 

 \section{Evaluation}

In this section we present an evaluation of our explainable surrogate model in the Atari domain. 
Specifically, we seek to identify the extent to which our surrogate model is able to successfully learn to approximate the behavior and underlying decision making of a target RL agent. 

\subsection{Implementation Details}

In terms of evaluation domain we chose a suite of Atari games, see Table \ref{results-table} for a list of these games. 
The first set of these games have simplified sprites and animations, which adhere to the implicit assumptions in Algorithm 1.
In particular, the sprites should not have any perspective, and the animation must be either alternating sprites (such as the fish in \textit{Seaquest}) or no animation (such as the ball in \textit{Breakout}).
These games should represent ideal evaluation domains for our approach.
Additionally, we chose two games with more complicated sprites (\textit{Boxing} and \textit{DemonAttack}), in order to demonstrate how our approach performs when these sprite assumptions are broken.
 
We choose to use OpenAI Gym \cite{brockman2016openai} for the environment implementations of the Atari games (which uses the Arcade Learning Environment \cite{ALE}), and the Stable Baselines \cite{stable-baselines} implementation of Rainbow \cite{hessel2017rainbow}, and frame skip.
We examine the cases of noop starts and sticky actions combined with noop starts.
We train the Rainbow agent for 1,000,000 timesteps on each game. 
We used the same parameters as in \cite{hessel2017rainbow}. 
We limit our noop starts to be [0, 29] inclusive, and for every 4 frames obtained from ALE, we use max(frame 3, frame 4) as an observation.
Since Rainbow uses a downsampled version of the pixel observations from the Atari games, we have downsampled the pixel representation provided by the Arcade Learning Environment from $(210, 160)$ to $(105, 80)$, in order to better represent the input that Rainbow is using while still minimising sprite distortion due to downsampling. 
For our surrogate models, we have chosen a decision tree \cite{steinberg2009cart} for its performance and high interpretability.
We have utilized Scikit-learn \cite{scikit-learn} for a decision tree implementation, using their default hyperparameters, which allows for easy adoption and extension of this work.

We require some method to understand what features are responsible for a given prediction deviating from the mean prediction to compare to our explanations.
Shapley values \cite{shapely} fit this requirement. 
We can consider the classification problem as a game, with the players as the features. 
If we assume the fairness axioms described in \cite{shapely}, then we have that the unique solution to a distribution of payout for each player is given by the Shapley value. 
SHAP \cite{lundberg2016unexpected} was created as a way to efficiently estimate Shapley values for individual predictions, and TreeSHAP \cite{lundberg2018consistent} gives a method for exactly calculating the Shapley values for tree based models in polynomial time. 
Thus, TreeSHAP is appropriate to obtain the exact Shapley values for our surrogate model, given we use a decision tree \cite{steinberg2009cart}.

\subsection{Experiments}
 
We chose to focus on the problem of demonstrating that our interpretable surrogate model accurately reflects the behavior and decision making of the RL agent. 
This is crucial, as while other approaches have demonstrated that even inaccurate explanations increase user trust \cite{ehsan2019automated,madumal2019explainable}, an accurate explanation is required for many domains and applications.
As such, we designed our experiments to answer the following research questions:

\begin{enumerate}
    \item Given some agent and environment, can a surrogate model be trained that is highly accurate?
    \item Does a surrogate model overfit to the trajectories it was trained on, or can it generalize to trajectories it hasn't been trained on?
    \item To what extent does the model capture the underlying decision making of the agent?
\end{enumerate}

\begin{table*}[t]
  \label{results-table}
  \centering
  \begin{tabular}{lllll}
    \toprule
    \cmidrule(r){1-5}
	  Game     & Accuracy (\%)     & Cross Entropy & Accuracy (\%) (Sticky)    & Cross Entropy (Sticky) \\
    \midrule
    Atlantis    & $89.5 \pm 0.129$ & $3.32 \pm 0.027$  & $81.6 \pm 0.209$ & $6.21 \pm 0.0740$   \\ 
	Gopher   & $93.51 \pm 0.056$ & $2.242 \pm 0.019$ & $84.6 \pm 0.178$ & $5.31 \pm 0.0580$ \\ 
    Breakout & $92.285 \pm 0.283$ & $2.66 \pm 0.097$ & $85.6 \pm 0.358$ & $4.93 \pm 0.122$   \\ 
    Star Gunner  & $87.477 \pm 0.309$ & $4.223 \pm 0.118$ & $73.9 \pm 0.345$ & $9.02 \pm 0.119$   \\ 
    Pong    & $90.271 \pm 0.027$ & $3.306 \pm 0.011$  & $82.369 \pm 0.169$ & $6.076 \pm 0.058$ \\ 
    Qbert   & $97.525 \pm 0.132$ & $0.524 \pm 0.025$ & $91.8 \pm 0.327$ & $2.83 \pm 0.117$  \\ 
    Space Invaders & $78.94 \pm 0.404$ & $7.047 \pm 0.128$  & $78.0 \pm 0.260$ & $7.50 \pm 0.0930$ \\ 
    Seaquest     & $93.2 \pm 0.234$ & $0.675 \pm 0.239$ & $66.2 \pm 0.120$ & $11.7 \pm 0.0420$  \\ 
    Ms Pacman   & $97.977 \pm 0.087$ & $0.47 \pm 0.03$ & $88.7 \pm 0.199$ & $3.79 \pm 0.0610$  \\ 
    Kung Fu Master  & $98.677 \pm 0.085$ & $0.373 \pm 0.035$  & $88.1 \pm 0.125$ & $4.10 \pm 0.0410$ \\ 
    Asterix  & $95.7 \pm 0.190$ & $0.137 \pm 0.007$   & $91.2 \pm 0.362$ & $2.85 \pm 0.124$ \\ 
    \hline
    Demon Attack  & $87.692 \pm 0.163$ & $4.251 \pm 0.056$  & $79.6 \pm 0.237$ & $7.03 \pm 0.082$ \\ 
    Boxing  & $74.11 \pm 0.169$ & $8.872 \pm 0.058$ & $77.6 \pm 0.249$ & $7.73 \pm 0.0870$ \\ 
    \bottomrule
  \end{tabular}
    \caption{Surrogate model accuracy and cross entropy. A line separates those games that match our sprite assumptions and those that do not.}
\end{table*}

In order to answer (1), we obtain 25 sampled trajectories using 0, 1, ..., 24 noop actions at the start of the episode, following the procedure outlined above, with and without sticky actions \cite{machado2017revisiting} with probability 0.25.
We train the surrogate in a supervised fashion, and evaluate both the accuracy and cross entropy using a 5 fold split. 
The results can be found in Table \ref{results-table}.
This demonstrates that it is possible to learn a very accurate approximation for the agent, at least for interpolation in between trajectories.
The usage of sticky actions causes much more variance in the trajectories \cite{machado2017revisiting}, causing greater diversity in the training data, likely being the cause of the performance reduction.
However, even in the sticky actions case, the resulting surrogate models are still very powerful despite being simple decision trees. 
We also note that for games like \textit{Boxing}, where the sprite animation is complicated due to the moving hand creating a variety of different sprites, there is a huge amount of sparsity in the data for a given state. 
However, despite this complexity and the fact that Boxing violated our sprite assumptions, the resulting surrogate model is still better than random chance.

\begin{table*}[t]

  \label{oob-table}
  \centering
  \begin{tabular}{lllll}
    \toprule
    \cmidrule(r){1-5}
	Game     & Accuracy (\%)     & Cross Entropy & Accuracy (\%) (Sticky)    & Cross Entropy (Sticky) \\
    \midrule
    Atlantis    & $96.2 \pm 0.791$ & $0.272 \pm 0.0252$   & $90.03 \pm 7.08$ & $3.29 \pm 2.45$ \\ 
    Gopher    & $98.1 \pm 0.115$ & $0.205 \pm 0.0159$  & $83.4 \pm 0.952$ & $5.71 \pm 0.338$ \\ 
    Breakout & $98.2 \pm 0.537$ & $0.158 \pm 0.00567$  & $96.33 \pm 2.04$ & $1.27 \pm 0.705$ \\ 
    Star Gunner & $95.0 \pm 0.971$ & $0.150 \pm 0.0207$  & $82.4 \pm 6.64$ & $6.02 \pm 2.29$ \\ 
    Pong     & $90.0 \pm 0.372 $  & $0.330 \pm 0.00671$  & $83.6 \pm 0.709$ & $5.64 \pm 0.240$ \\ 
    Qbert     & $99.1 \pm 0.319 $ & $0.0282 \pm 0.00505$ & $97.19 \pm 1.29$ & $0.703 \pm 0.481$  \\ 
    Space Invaders & $99.5 \pm 0.00265$ & $0.118 \pm 0.0125$ & $65.3 \pm 5.34$ & $11.7 \pm 1.88$ \\ 
    Seaquest     & $99.0 \pm 0.185$ & $0.121 \pm 0.00815$  & $72.6 \pm 4.06$ & $9.47 \pm 1.40$ \\ 
    Ms Pacman     & $97.4 \pm 0.00694$ & $0.0620 \pm 0.00909$  & $98.5 \pm 0.443$ & $0.121 \pm 0.023$ \\ 
    Kung Fu Master      & $99.9 \pm 0.0361$ & $0.324 \pm 0.00345$  & $85.5 \pm 1.74$ & $4.93 \pm 0.581$ \\ 
    Asterix     & $95.7 \pm 0.19$ & $0.137 \pm 0.007$  & $66.9 \pm 6.86$ & $10.9 \pm 2.33$ \\ 
    \hline
    Demon Attack     & $67.8 \pm 6.38$ & $0.964 \pm 0.139$  & $64.6 \pm 2.13$ & $12.2 \pm 0.737$ \\ 
	  Boxing     & $58.0 \pm 2.33$ & $1.33 \pm 0.08356$  & $76.2 \pm 0.547$ & $8.20 \pm 0.189$ \\ 
    \bottomrule
  \end{tabular}
    \caption{Surrogate model accuracy and cross entropy on unseen trajectories.}
\end{table*}

Now, the next question is if this surrogate model can generalize to trajectories that it has never seen before. 
This is equivalent to answering (2).
We consider this problem solely in the domain of noop starts, without loss of generality.
Note that the agent attempts to maximize its return over trajectories where the starting state is sampled from choosing 0, 2, ..., 29 noop starts.
We have trained the model on the trajectories using 0, 2, ..., 24 noop starts, leaving the trajectories generated by 25, 26, ..., 29 noop starts as a test set. 
As such, we obtain the 5 trajectories aforementioned, using the method outlined in section 3.2. We then use this dataset as a test set for the surrogate model. 
The results can be found in Table \ref{oob-table}. 
We see that the performance of the surrogate model is still strong, in some cases stronger than in the training set.  
This indicates that the model has learned a good, general model of the behavior of the agent.
The consistently strong accuracy results also stand out in comparison to other XAI approaches \cite{madumal2019explainable,ehsan2019automated}.
There is a significant degradation in performance over the unseen trajectories for \textit{Boxing}, likely due to the lack of features that describe objects relative to one another in the feature space, greatly restricting the ability for the model to learn location invariant features. 
Said features appear to be important to the policy learned for the non-sticky action case.

\begin{table}[t]
  \label{permute-table}
  \centering
  \begin{tabular}{ll}
    \toprule
    \cmidrule(r){1-2}
    Game     &  Agent action changed \%  \\
    \midrule
    Atlantis    & 54.8     \\
    Gopher    &  46.8  \\
    Breakout & 92.5  \\
    Star Gunner & 69.2   \\
    Pong     & 92.8   \\
    Qbert     & 74.1  \\
    Space Invaders & 70.0 \\
    Seaquest     & 76.8    \\
    Ms Pacman     & 94.6   \\
    Kung Fu Master    & 73.6    \\
    Asterix    & 40.6    \\
    Demon Attack & 62.7     \\
    Boxing & 86.0   \\
    \bottomrule
  \end{tabular}
    \caption{Probability of agent changing action given permuted state.}
\end{table}

Our third research question asks whether or not the learned decision making of the model accurately approximates the underlying decision making of the agent.
This is important, as it's possible our surrogate model is getting the right answer for the wrong reason, or a type III error. 
But we cannot directly compare the decision making as that would require matching decision nodes and neural network weights. 
However, if it is the case that the surrogate model has accurately approximated the Rl agent's policy in symbolic features, than these symbolic features should be integral to the RL agent's policy. 
As such, we show that a permutation of the symbolic features important to the surrogate model's decision making also results in a significant change in agent behavior. 

We use the surrogate model to produce adversarial examples \cite{goodfellow2014explaining}. 
Instead of a single decision tree, we employ a tree ensemble (LightGBM \cite{NIPS2017_6907} with default hyperparameters) to allow for many possible candidate splits to search through in order to generate said adversarial examples.
We use the degree of change in the prediction of the surrogate model as a metric for producing adversarial examples, as such a single decision tree would be insufficient.
To begin with, we take all sprites in the state, and rank them according to the maximal Shapley value of each of their features for a decision tree's predicted action. 
We then consider the sprites in the top 10\% of this list. For each decision tree in our surrogate model, we do a 3-ply search over the splits, obtaining eight different subsets of the dataset. 
We sample one state from each subset. 
For each sampled state, we swap the feature values corresponding to the top 10\% of sprites with the feature values in the sampled state. 
This gives us 800 possible adversarial examples, as we employ a surrogate model of 100 decision trees. 
We take the example which minimizes the original prediction of the surrogate model to be our adversarial example.
This is essentially the permutation of the original state that most changed the predicted behavior from our surrogate model. 
At no point do we query the deep RL agent when producing these adversarial examples. 
If the agent is using the same features in its decision making, it follows that it's behavior should change when the surrogate model's prediction changes.

We collected 200 (state, action) pair permutations from the trajectory with 24 noop starts, and considered only the (state, action) pairs.
The results in terms of the percentage of instances where the agent changed its action from the original compared to the permuted state can be found in Table \ref{permute-table}. 

We can see that in general, our adversarial examples cause the agent to change its predicted action, with an average probability of ~72\% across all games. 
Due to the nature of these comparisons, we consider them similar to correlation, which would imply a strong average alignment in decision making between the surrogate and agent.
This is surprising given how much smaller and simpler our surrogate is compared to Rainbow, even in the case of using a random forest in comparison to a single decision tree. 
The alignment remained strong in the more complex games where our surrogate's action prediction accuracy was lower. 
This indicates that even in the more complex games our surrogate is still identifying the important features on which the agent makes its decisions. 
We do not anticipate that this was due to the use of a random forest instead of a single decision tree, as the decision tree performed roughly equivalently to the random forest across the first two evaluations in our experiments. 
We note that Rainbow is a limiting factor here, as we cannot measure the confidence change in the action, which would give us a more nuanced measure of impact.

\begin{figure}[h]
  \centering
  \includegraphics[width=7.3cm]{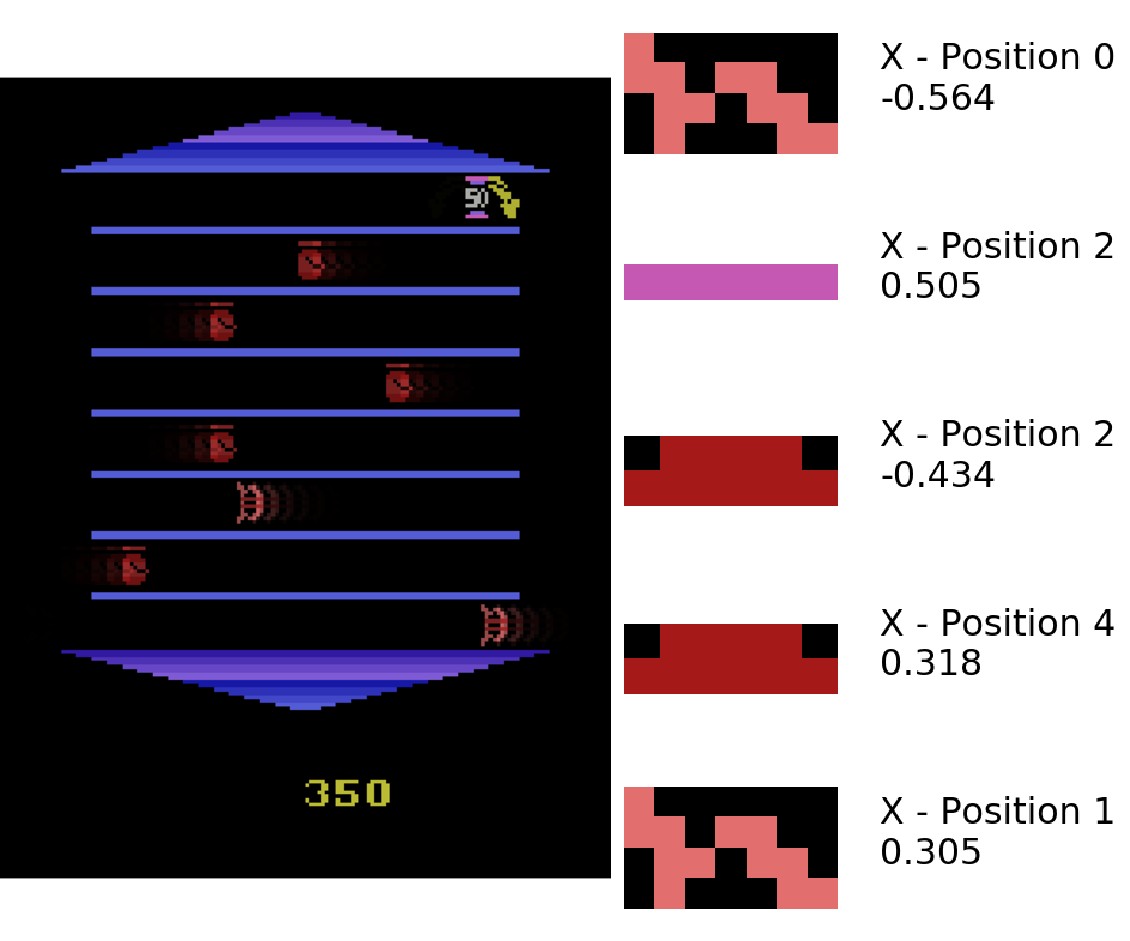}
  
  \vspace{0.5cm}
  
  \includegraphics[width=7.3cm]{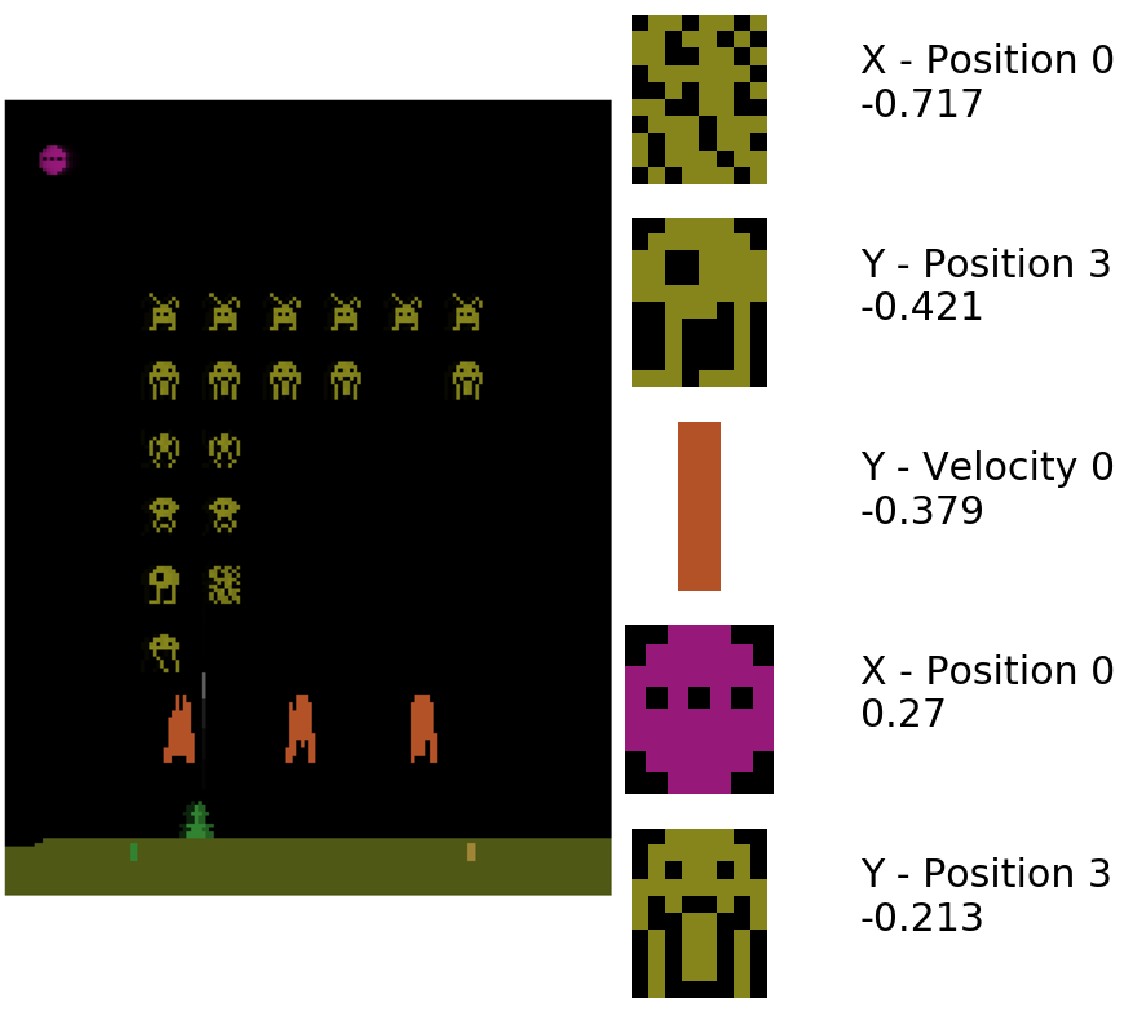}
  
  \caption{
  A pictorial representation of the highest absolute Shapley values given a specific game state. 
  The left frame occurs 1 time step before the right frame. 
  The 5 sprites on the right are ordered by their Shapley value, and additional information (which coordinate, position or velocity, and which sprite (ordered from top left to bottom right)).
  }
  \label{shap}
\end{figure}

\begin{figure}[h]
  
  \hspace{-.5cm}\includegraphics[width=9.5cm]{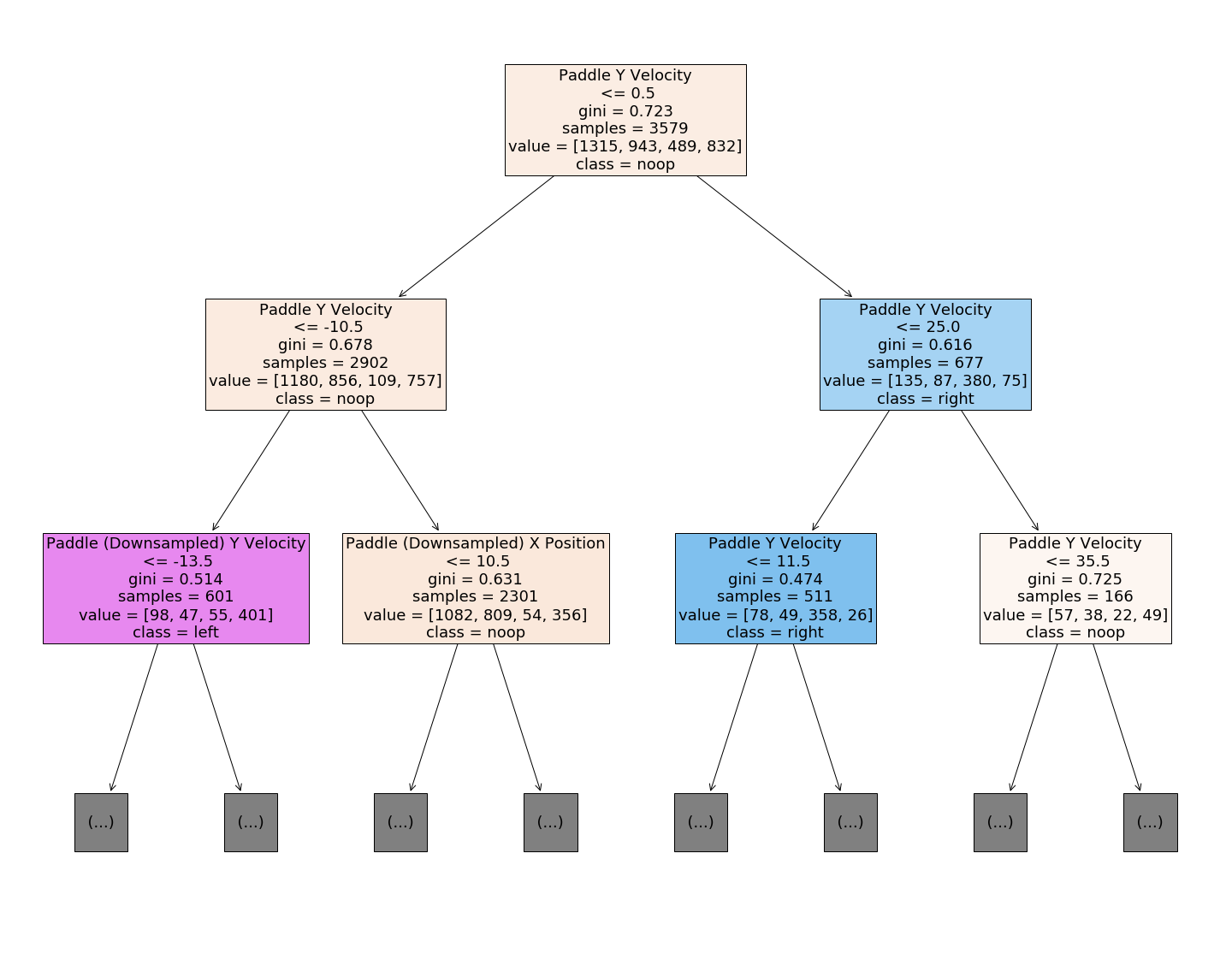}
  
  \caption{
  A visualization of the first 3 levels of the decision tree classifier learned on the training data from Breakout. 
  The colors correspond to the chosen label; tan corresponds to class 0 (noop), blue corresponds to class 2 (move right) and pink corresponds to class 3 (move left).
  }
  \label{dt}
\end{figure}

We have shown thus far that the model is a strong surrogate of the agent, both for interpolation of trajectories previously seen and on trajectories not explicitly seen before, and that the decision making of both the agent and model is well-aligned. 
However, this does not necessarily mean that the features learned by the surrogate model also match human intuition.
In Figure \ref{shap} we visualize the sprites with the highest Shapley values in the selected frames for two of the Atari games. 
We can see that the player sprite, along with the sprites the player would interact with impacts the deviation of the given predicted action distribution from the mean predicted action distribution, with little regard to other objects of the scene.
This agrees with our intuition for these games, in terms of which sprites are of most importance.
Additionally, in figure \ref{dt}, we visualize the first three levels of the decision tree obtained from \textit{Breakout}, all of which focus on paddle orientation.
There is a huge emphasis on paddle orientation and velocity that agrees with human intuition, such as continuing to move left when already moving left, and likewise with moving right. 
Additionally, opting to do noops after coming in contact with the ball is a common strategy observed with Breakout agents trained by DQN, which is emphasized by the Paddle X position node. 
One oddity is that the ball touching the paddle is recognized as a unique sprite compared to the ball or paddle individually. 
This is due to the simple sprite finding technique utilized, and the fact that both ball and paddle are the same color. 
However, represented visually, and with some knowledge of \textit{Breakout}, the sprite still reads clearly.

\section{Limitations and Future Work}

We note a number of limitations and corresponding areas of future work.
From the results with \textit{Demon Attack} and \textit{Boxing}, we can identify that our approach at times underperforms in more complex games. 
We anticipate that this is partially due to our naive sprite detection approach.
In future work we plan to draw on techniques like Blob Detection \cite{blob} and image segmentation \cite{felzenszwalb2004efficient} to address this.

Atari is a relatively simple domain. 
While this is appropriate for an initial exploration of this approach, given it's popularity for deep RL research, it is far less complex than many real world domains. 
While we drew upon intuitions unique to video games in this work, these assumptions will not hold in other domains. 
We anticipate that these other domains, particularly those reliant on natural language, will require their own non-trivial state transformations. 

We do not include a human subject study, instead focusing this paper on demonstrating that our method shows strong accuracy with regards to an RL agent compared to prior work \cite{madumal2019explainable,ehsan2019automated}.
In particular, interpretability, as discussed in this paper, relies on the intrinsic interpretability of decision trees and Shapley values.
It's uncertain whether or not these decision trees are practically interpretable with human subjects, which will be explored in further work or how they might be employed to produce explanations for those without technical knowledge.
We intend to delve into the specific impact of our surrogate model on human experience in future work.

\section{Conclusion}
In this paper, we have presented a simple, yet surprisingly effective approach to train an accurate surrogate model. 
We proved that our interpretable representation is a bijection of the original state space. 
Through experimentation we demonstrated that the surrogate model learned to accurately approximate the behavior of Rainbow on a set of Atari games.
This indicates that the approach is an appropriate one to add interpretability to complex, black box agents in simple domains.

\section*{Acknowledgements}

We acknowledge the support of the Natural Sciences and Engineering Research Council of Canada (NSERC) and the Alberta Machine Intelligence Institute (Amii).

\section*{Ethics Statement}

We introduce an Explainable AI approach, which has the potential to offer broad impacts in terms of how and where Reinforcement Learning can be applied. 
Our approach is simple, but surprisingly powerful. 
This is beneficial, as it indicates it will be easier for others to adapt this approach to their needs. 
However, because our approach only demonstrated strong performance on simple domains, there is the potential for negative impact if it is brought to more complex domains too soon. 
Without sufficient improvement, our approach applied to a complex RL agent may present inaccurate explanations that still increase user satisfaction and trust \cite{madumal2019explainable}.
This could be devastating in high-risk scenarios, for example doctors trusting faulty diagnoses \cite{holzinger2017need}.
We therefore highly caution against applying this approach broadly until it has been improved to function in more complex domains and suitable research into its impact on users has been completed.

\bibliography{aaai21}

\begin{thebibliography}{40}
\providecommand{\natexlab}[1]{#1}
\providecommand{\url}[1]{\texttt{#1}}
\providecommand{\urlprefix}{URL }
\expandafter\ifx\csname urlstyle\endcsname\relax
  \providecommand{\doi}[1]{doi:\discretionary{}{}{}#1}\else
  \providecommand{\doi}{doi:\discretionary{}{}{}\begingroup
  \urlstyle{rm}\Url}\fi

\bibitem[{EUd(2018)}]{EUdataregulations2018}
 2018.
\newblock 2018 reform of EU data protection rules.
\newblock
  \urlprefix\url{https://ec.europa.eu/commission/sites/beta-political/files/data-protection-factsheet-changes_en.pdf}.

\bibitem[{Bellemare et~al.(2013)Bellemare, Naddaf, Veness, and Bowling}]{ALE}
Bellemare, M.~G.; Naddaf, Y.; Veness, J.; and Bowling, M. 2013.
\newblock The Arcade Learning Environment: An Evaluation Platform for General
  Agents.
\newblock \emph{Journal of Artificial Intelligence Research} 47: 253–279.
\newblock ISSN 1076-9757.
\newblock \doi{10.1613/jair.3912}.
\newblock \urlprefix\url{http://dx.doi.org/10.1613/jair.3912}.

\bibitem[{Brockman et~al.(2016)Brockman, Cheung, Pettersson, Schneider,
  Schulman, Tang, and Zaremba}]{brockman2016openai}
Brockman, G.; Cheung, V.; Pettersson, L.; Schneider, J.; Schulman, J.; Tang,
  J.; and Zaremba, W. 2016.
\newblock OpenAI Gym.
\newblock \urlprefix\url{http://arxiv.org/abs/1606.01540}.
\newblock Cite arxiv:1606.01540.

\bibitem[{{Danker} and {Rosenfeld}(1981)}]{blob}
{Danker}, A.~J.; and {Rosenfeld}, A. 1981.
\newblock Blob Detection by Relaxation.
\newblock \emph{IEEE Transactions on Pattern Analysis and Machine Intelligence}
  PAMI-3(1): 79--92.

\bibitem[{Doshi-Velez and Kim(2017)}]{doshivelez2017rigorous}
Doshi-Velez, F.; and Kim, B. 2017.
\newblock Towards A Rigorous Science of Interpretable Machine Learning.
\newblock \url{https://arxiv.org/abs/1702.08608}.

\bibitem[{Ehsan et~al.(2017)Ehsan, Harrison, Chan, and
  Riedl}]{ehsan2017rationalization}
Ehsan, U.; Harrison, B.; Chan, L.; and Riedl, M.~O. 2017.
\newblock Rationalization: A Neural Machine Translation Approach to Generating
  Natural Language Explanations.
\newblock \url{https://arxiv.org/pdf/1702.07826.pdf}.

\bibitem[{Ehsan et~al.(2019)Ehsan, Tambwekar, Chan, Harrison, and
  Riedl}]{ehsan2019automated}
Ehsan, U.; Tambwekar, P.; Chan, L.; Harrison, B.; and Riedl, M. 2019.
\newblock Automated Rationale Generation: A Technique for Explainable AI and
  its Effects on Human Perceptions.
\newblock \url{https://arxiv.org/pdf/1901.03729.pdf}.

\bibitem[{Felzenszwalb and Huttenlocher(2004)}]{felzenszwalb2004efficient}
Felzenszwalb, P.~F.; and Huttenlocher, D.~P. 2004.
\newblock Efficient graph-based image segmentation.
\newblock \emph{International journal of computer vision} 59(2): 167--181.

\bibitem[{Goebel et~al.(2018)Goebel, Chander, Holzinger, Lecue, Akata, Stumpf,
  Kieseberg, and Holzinger}]{goebel2018explainable}
Goebel, R.; Chander, A.; Holzinger, K.; Lecue, F.; Akata, Z.; Stumpf, S.;
  Kieseberg, P.; and Holzinger, A. 2018.
\newblock Explainable AI: the new 42?
\newblock In \emph{International Cross-Domain Conference for Machine Learning
  and Knowledge Extraction}, 295--303. Springer.

\bibitem[{Goodfellow, Shlens, and Szegedy(2014)}]{goodfellow2014explaining}
Goodfellow, I.~J.; Shlens, J.; and Szegedy, C. 2014.
\newblock Explaining and harnessing adversarial examples.
\newblock \emph{arXiv preprint arXiv:1412.6572} .

\bibitem[{Greydanus et~al.(2018)Greydanus, Koul, Dodge, and
  Fern}]{greydanus2018visualizing}
Greydanus, S.; Koul, A.; Dodge, J.; and Fern, A. 2018.
\newblock Visualizing and understanding atari agents.
\newblock In \emph{International Conference on Machine Learning}, 1792--1801.

\bibitem[{Guidotti et~al.(2018)Guidotti, Monreale, Ruggieri, Turini, Giannotti,
  and Pedreschi}]{10.1145/3236009}
Guidotti, R.; Monreale, A.; Ruggieri, S.; Turini, F.; Giannotti, F.; and
  Pedreschi, D. 2018.
\newblock A Survey of Methods for Explaining Black Box Models.
\newblock \emph{ACM Comput. Surv.} 51(5).
\newblock ISSN 0360-0300.
\newblock \doi{10.1145/3236009}.
\newblock \urlprefix\url{https://doi.org/10.1145/3236009}.

\bibitem[{Hayes and Shah(2017)}]{hayes2017improving}
Hayes, B.; and Shah, J.~A. 2017.
\newblock Improving robot controller transparency through autonomous policy
  explanation.
\newblock In \emph{2017 12th ACM/IEEE International Conference on Human-Robot
  Interaction (HRI}, 303--312. IEEE.

\bibitem[{Hessel et~al.(2017)Hessel, Modayil, van Hasselt, Schaul, Ostrovski,
  Dabney, Horgan, Piot, Azar, and Silver}]{hessel2017rainbow}
Hessel, M.; Modayil, J.; van Hasselt, H.; Schaul, T.; Ostrovski, G.; Dabney,
  W.; Horgan, D.; Piot, B.; Azar, M.; and Silver, D. 2017.
\newblock Rainbow: Combining Improvements in Deep Reinforcement Learning.
\newblock \url{https://arxiv.org/pdf/1710.02298.pdf}.

\bibitem[{Hill et~al.(2018)Hill, Raffin, Ernestus, Gleave, Kanervisto, Traore,
  Dhariwal, Hesse, Klimov, Nichol, Plappert, Radford, Schulman, Sidor, and
  Wu}]{stable-baselines}
Hill, A.; Raffin, A.; Ernestus, M.; Gleave, A.; Kanervisto, A.; Traore, R.;
  Dhariwal, P.; Hesse, C.; Klimov, O.; Nichol, A.; Plappert, M.; Radford, A.;
  Schulman, J.; Sidor, S.; and Wu, Y. 2018.
\newblock Stable Baselines.
\newblock \url{https://github.com/hill-a/stable-baselines}.

\bibitem[{Holzinger et~al.(2017)Holzinger, Biemann, Pattichis, and
  Kell}]{holzinger2017need}
Holzinger, A.; Biemann, C.; Pattichis, C.~S.; and Kell, D.~B. 2017.
\newblock What do we need to build explainable AI systems for the medical
  domain?
\newblock \url{https://arxiv.org/pdf/1712.09923.pdf}.

\bibitem[{Hussein et~al.(2017)Hussein, Gaber, Elyan, and Jayne}]{imitation}
Hussein, A.; Gaber, M.~M.; Elyan, E.; and Jayne, C. 2017.
\newblock Imitation Learning: A Survey of Learning Methods.
\newblock \emph{ACM Comput. Surv.} 50(2).
\newblock ISSN 0360-0300.
\newblock \doi{10.1145/3054912}.
\newblock \urlprefix\url{https://doi.org/10.1145/3054912}.

\bibitem[{Juozapaitis et~al.(2019)Juozapaitis, Koul, Fern, Erwig, and
  Doshi-Velez}]{explainablerewarddecomp}
Juozapaitis, Z.; Koul, A.; Fern, A.; Erwig, M.; and Doshi-Velez, F. 2019.
\newblock Explainable Reinforcement Learning via Reward Decomposition.
\newblock \emph{Proceedings of the Twenty-Sixth International Joint Conference
  on Artificial Intelligence, {IJCAI-19}} .

\bibitem[{Ke et~al.(2017)Ke, Meng, Finley, Wang, Chen, Ma, Ye, and
  Liu}]{NIPS2017_6907}
Ke, G.; Meng, Q.; Finley, T.; Wang, T.; Chen, W.; Ma, W.; Ye, Q.; and Liu,
  T.-Y. 2017.
\newblock LightGBM: A Highly Efficient Gradient Boosting Decision Tree.
\newblock In Guyon, I.; Luxburg, U.~V.; Bengio, S.; Wallach, H.; Fergus, R.;
  Vishwanathan, S.; and Garnett, R., eds., \emph{Advances in Neural Information
  Processing Systems 30}, 3146--3154. Curran Associates, Inc.

\bibitem[{Lundberg and Lee(2016)}]{lundberg2016unexpected}
Lundberg, S.; and Lee, S.-I. 2016.
\newblock An unexpected unity among methods for interpreting model predictions.
\newblock \url{https://arxiv.org/pdf/1611.07478.pdf}.

\bibitem[{Lundberg, Erion, and Lee(2018)}]{lundberg2018consistent}
Lundberg, S.~M.; Erion, G.~G.; and Lee, S.-I. 2018.
\newblock Consistent Individualized Feature Attribution for Tree Ensembles.
\newblock \url{https://arxiv.org/abs/1802.03888}.

\bibitem[{Maaten and Hinton(2008)}]{maaten2008visualizing}
Maaten, L. v.~d.; and Hinton, G. 2008.
\newblock Visualizing data using t-SNE.
\newblock \emph{Journal of machine learning research} 9(Nov): 2579--2605.

\bibitem[{Machado et~al.(2017)Machado, Bellemare, Talvitie, Veness, Hausknecht,
  and Bowling}]{machado2017revisiting}
Machado, M.~C.; Bellemare, M.~G.; Talvitie, E.; Veness, J.; Hausknecht, M.; and
  Bowling, M. 2017.
\newblock Revisiting the Arcade Learning Environment: Evaluation Protocols and
  Open Problems for General Agents.
\newblock \url{https://arxiv.org/abs/1709.06009}.

\bibitem[{Madumal et~al.(2019)Madumal, Miller, Sonenberg, and
  Vetere}]{madumal2019explainable}
Madumal, P.; Miller, T.; Sonenberg, L.; and Vetere, F. 2019.
\newblock Explainable Reinforcement Learning Through a Causal Lens.
\newblock \url{https://arxiv.org/abs/1905.10958}.

\bibitem[{Mnih et~al.(2013)Mnih, Kavukcuoglu, Silver, Graves, Antonoglou,
  Wierstra, and Riedmiller}]{mnih2013playing}
Mnih, V.; Kavukcuoglu, K.; Silver, D.; Graves, A.; Antonoglou, I.; Wierstra,
  D.; and Riedmiller, M. 2013.
\newblock Playing Atari with Deep Reinforcement Learning.
\newblock \url{https://www.cs.toronto.edu/~vmnih/docs/dqn.pdf}.

\bibitem[{Mnih et~al.(2015)Mnih, Kavukcuoglu, Silver, Rusu, Veness, Bellemare,
  Graves, Riedmiller, Fidjeland, Ostrovski et~al.}]{mnih2015human}
Mnih, V.; Kavukcuoglu, K.; Silver, D.; Rusu, A.~A.; Veness, J.; Bellemare,
  M.~G.; Graves, A.; Riedmiller, M.; Fidjeland, A.~K.; Ostrovski, G.; et~al.
  2015.
\newblock Human-level control through deep reinforcement learning.
\newblock \emph{Nature} 518(7540): 529--533.

\bibitem[{Molnar(2019)}]{molnar2019}
Molnar, C. 2019.
\newblock \emph{Interpretable Machine Learning}.
\newblock Accessed: 2019-04-12.

\bibitem[{Pedregosa et~al.(2011)Pedregosa, Varoquaux, Gramfort, Michel,
  Thirion, Grisel, Blondel, Prettenhofer, Weiss, Dubourg, Vanderplas, Passos,
  Cournapeau, Brucher, Perrot, and Duchesnay}]{scikit-learn}
Pedregosa, F.; Varoquaux, G.; Gramfort, A.; Michel, V.; Thirion, B.; Grisel,
  O.; Blondel, M.; Prettenhofer, P.; Weiss, R.; Dubourg, V.; Vanderplas, J.;
  Passos, A.; Cournapeau, D.; Brucher, M.; Perrot, M.; and Duchesnay, E. 2011.
\newblock Scikit-learn: Machine Learning in {P}ython.
\newblock \emph{Journal of Machine Learning Research} 12: 2825--2830.

\bibitem[{Ribeiro, Singh, and Guestrin(2016)}]{ribeiro2016modelagnostic}
Ribeiro, M.~T.; Singh, S.; and Guestrin, C. 2016.
\newblock Model-Agnostic Interpretability of Machine Learning.
\newblock \url{https://arxiv.org/pdf/1606.05386.pdf}.

\bibitem[{Rudin and Ustun(2018)}]{rudin2018optimized}
Rudin, C.; and Ustun, B. 2018.
\newblock Optimized scoring systems: Toward trust in machine learning for
  healthcare and criminal justice.
\newblock \emph{Interfaces} 48(5): 449--466.

\bibitem[{Sado et~al.(2020)Sado, Loo, Kerzel, and
  Wermter}]{sado2020explainable}
Sado, F.; Loo, C.~K.; Kerzel, M.; and Wermter, S. 2020.
\newblock Explainable Goal-Driven Agents and Robots -- A Comprehensive Review
  and New Framework.
\newblock \url{https://arxiv.org/abs/2004.09705}.

\bibitem[{Schaul et~al.(2015)Schaul, Quan, Antonoglou, and
  Silver}]{schaul2015prioritized}
Schaul, T.; Quan, J.; Antonoglou, I.; and Silver, D. 2015.
\newblock Prioritized Experience Replay.
\newblock \url{https://arxiv.org/abs/1511.05952}.

\bibitem[{Shalev-Shwartz, Shammah, and Shashua(2016)}]{shalevshwartz2016safe}
Shalev-Shwartz, S.; Shammah, S.; and Shashua, A. 2016.
\newblock Safe, Multi-Agent, Reinforcement Learning for Autonomous Driving.
\newblock \url{https://arxiv.org/pdf/1610.03295.pdf}.

\bibitem[{Shapley(1952)}]{shapely}
Shapley, L.~S. 1952.
\newblock A Value for n-Person Games.
\newblock \emph{RAND Corporation} 295.

\bibitem[{Simonyan, Vedaldi, and Zisserman(2013)}]{simonyan2013deep}
Simonyan, K.; Vedaldi, A.; and Zisserman, A. 2013.
\newblock Deep Inside Convolutional Networks: Visualising Image Classification
  Models and Saliency Maps.
\newblock \url{https://arxiv.org/abs/1312.6034}.

\bibitem[{Steinberg and Colla(2009)}]{steinberg2009cart}
Steinberg, D.; and Colla, P. 2009.
\newblock CART: classification and regression trees.
\newblock \emph{The top ten algorithms in data mining} 9: 179.

\bibitem[{Sutton and Barto(2018)}]{10.5555/3312046}
Sutton, R.~S.; and Barto, A.~G. 2018.
\newblock \emph{Reinforcement Learning: An Introduction}.
\newblock Cambridge, MA, USA: A Bradford Book.
\newblock ISBN 0262039249.

\bibitem[{van Hasselt, Guez, and Silver(2015)}]{hasselt2015deep}
van Hasselt, H.; Guez, A.; and Silver, D. 2015.
\newblock Deep Reinforcement Learning with Double Q-learning.
\newblock \url{https://arxiv.org/abs/1509.06461}.

\bibitem[{Wang et~al.(2015)Wang, Schaul, Hessel, van Hasselt, Lanctot, and
  de~Freitas}]{wang2015dueling}
Wang, Z.; Schaul, T.; Hessel, M.; van Hasselt, H.; Lanctot, M.; and de~Freitas,
  N. 2015.
\newblock Dueling Network Architectures for Deep Reinforcement Learning.
\newblock \url{https://arxiv.org/abs/1511.06581}.

\bibitem[{Watkins and Dayan(1992)}]{watkins}
Watkins, C.~J.; and Dayan, P. 1992.
\newblock Q-learning.
\newblock \emph{Machine learning} 8(3-4): 279--292.

\end{thebibliography}

\end{document}